\def\year{2019}\relax
\documentclass[letterpaper]{article} 
\usepackage{aaai19}  
\usepackage{times}  
\usepackage{helvet}  
\usepackage{courier}  
\usepackage{url}  
\usepackage{graphicx}  
\usepackage{amsmath,amssymb, amsthm}
\usepackage{textcomp}
\usepackage{bm}
\usepackage{adjustbox}

\usepackage{booktabs}
\usepackage{bm}
\usepackage{mathtools}
\usepackage{multirow}

\usepackage{tikz}
\usepackage{subcaption}

\usepackage{enumitem}
\begin{document}
%
\title{Deep Interest Evolution Network for Click-Through Rate Prediction}
\author{Guorui Zhou\thanks{Corresponding author is Guorui Zhou.}, Na Mou\thanks{This author is the one who did the really hard work for Online Testing. The source code is available at https://github.com/mouna99/dien.}, Ying Fan, Qi Pi, Weijie Bian,\\
	{\bf \Large Chang Zhou, Xiaoqiang Zhu \and Kun Gai}\\
	 Alibaba Inc, Beijing, China\\
	\{guorui.xgr, mouna.mn, fanying.fy, piqi.pq, weijie.bwj, ericzhou.zc, xiaoqiang.zxq, jingshi.gk\}@alibaba-inc.com
}
\maketitle
\begin{abstract}
Click-through rate~(CTR) prediction, whose goal is to estimate the probability of a user clicking on the item, has become one of the core tasks in the advertising system.
For CTR prediction model, it is necessary to capture the latent user interest behind the user behavior data.
Besides, considering the changing of the external environment and the internal cognition,  user interest evolves over time dynamically. 
There are several CTR prediction methods for interest modeling, while most of them regard the representation of behavior as the interest directly, and lack specially modeling for latent interest behind the concrete behavior. Moreover, little work considers the changing trend of the interest. 
In this paper, we propose a novel model, named Deep Interest Evolution Network~(DIEN), for CTR prediction. Specifically, we design interest extractor layer to capture temporal interests from history behavior sequence. At this layer,  
we introduce an auxiliary loss to supervise interest extracting at each step. 
As user interests are diverse, especially in the e-commerce system, we propose interest evolving layer to capture interest evolving process that is relative to the target item. At interest evolving layer, attention mechanism  is embedded into the sequential structure novelly, and the effects of relative interests are strengthened during interest evolution.   
In the experiments on both public and industrial datasets, DIEN significantly outperforms the state-of-the-art solutions. Notably, DIEN has been deployed in the display advertisement system of Taobao, and obtained 20.7\% improvement on CTR.  
\end{abstract}
\section{Introduction}
Cost per click~(CPC) billing is one of the commonest billing forms in the advertising system, where advertisers are charged for each click on their advertisement. In CPC advertising system, the performance of click-through rate~(CTR) prediction not only influences the final revenue of whole platform, but also impacts user experience and satisfaction. Modeling CTR prediction has drawn more and more attention from the communities of academia and industry.

In most non-searching e-commerce scenes, users do not express their current intention actively. Designing model to capture user's interests as well as their dynamics is the key to advance the performance of CTR prediction. 
Recently, many CTR models transform from traditional methodologies~\cite{friedman2001greedy,rendle2010factorization} to deep CTR models~\cite{guo2017deepfm,qu2016product,lian2018xdeepfm}. Most deep CTR models focus on capturing interaction between features from different fields and pay less attention to user interest representation. 
Deep Interest Network~(DIN)~\cite{zhou2018deep} emphasizes that user interests are diverse, it uses attention based model to capture relative interests to target item, and obtains adaptive interest representation.  
However, most interest models including DIN regard the behavior as the interest directly, while latent interest is hard to be fully reflected by explicit behavior. Previous methods neglect to dig the true user interest behind behavior. Moreover, user interest keeps evolving, capturing the dynamic of interest is important for interest representation.\par
Based on all these observations, we propose Deep Interest Evolution Network~(DIEN) to improve the performance of CTR prediction. 
There are two key modules in DIEN, one is for extracting latent temporal interests from explicit user behaviors, 
and the other one is for modeling interest evolving process. 
Proper interest representation is the footstone of interest evolving model. At interest extractor layer, 
DIEN chooses GRU~\cite{chung2014empirical} to model the dependency between behaviors. 
Following the principle that interest leads to the consecutive behavior directly, we propose auxiliary loss 
which uses the next behavior to supervise the learning of current hidden state. We call these hidden states with extra supervision as interest states.
These extra supervision information helps to capture more semantic meaning for interest representation and push hidden states of GRU to represent interests effectively. 
Moreover, user interests are diverse, which leads to interest drifting phenomenon: user's intentions can be very different in adjacent visitings, 
and one behavior of a user may depend on the behavior that takes long time ago. 
Each interest has its own evolution track. 
Meanwhile, the click actions of one user on different target items are effected by different parts of interests. 
At interest evolving layer, we model the interest evolving trajectory that is relative to target item.
Based on the interest sequence obtained from interest extractor layer, we design GRU with attentional update gate~(AUGRU). Using interest state and target item to compute relevance, AUGRU strengthens relative interests' influence on interest evolution, while weakens irrelative interests' effect that results from interest drifting. With the introduction of attentional mechanism into update gate, AUGRU can lead to the specific interest evolving processes for different target items. 
The main contributions of DIEN are as following:
\begin{itemize}
	\item{We focus on interest evolving phenomenon in e-commerce system, and propose a new structure of network to model interest evolving process. The model for interest evolution leads to more expressive interest representation and more precise CTR prediction.}
	\item{Different from taking behaviors as interests directly, we specially design interest extractor layer.
		Pointing at the problem that hidden state of GRU is less targeted for interest representation, we propose one auxiliary loss. Auxiliary loss uses consecutive behavior to supervise the learning of hidden state at each step. which makes hidden state expressive enough to represent latent interest.}
	\item{We design interest evolving layer novelly, where GPU with attentional update gate~(AUGRU) strengthens the effect from relevant interests to target item and overcomes the inference from interest drifting.}
\end{itemize}
In the experiments on both public and industrial datasets, DIEN significantly outperforms the state-of-the-art solutions. It is notable that DIEN has been deployed in Taobao display advertisement system and obtains significant improvement under various metrics.
\section{Related Work}
By virtue of the strong ability of deep learning on feature representation and combination, recent CTR models transform from traditional linear or nonlinear models~\cite{friedman2001greedy,rendle2010factorization} to deep models.  Most deep models follow the structure of Embedding and Multi-ayer Perceptron~(MLP)~\cite{zhou2018deep}. 
Based on this basic paradigm, more and more models pay attention to the interaction between features: Both Wide \& Deep~\cite{cheng2016wide} and deep FM~\cite{guo2017deepfm} combine low-order and high-order features to improve the power of expression; PNN~\cite{qu2016product} proposes a product layer to capture interactive patterns between interfield categories. 
However, these methods can not reflect the interest behind data clearly. DIN~\cite{zhou2018deep} introduces the mechanism of attention to activate the historical behaviors w.r.t. given target item locally, and captures the diversity characteristic of user interests successfully. However, DIN is weak in capturing the dependencies between sequential behaviors.

In many application domains, user-item interactions can be recorded over time. A number of recent studies show that this information can be used to build richer individual user
models and discover additional behavioral patterns.
In recommendation system, TDSSM~\cite{song2016multi} jointly optimizes long-term and short-term user interests to improve the recommendation quality; DREAM~\cite{yu2016dynamic} uses the structure of recurrent neural network~(RNN) to investigate the dynamic representation of each user and the global sequential behaviors of item-purchase history. \citeauthor{he2016ups}~\shortcite{he2016ups} build visually-aware recommender system which surfaces products that more closely match users' and communities' evolving interests.  \citeauthor{zhang2014sequential}~\shortcite{zhang2014sequential} 
measures users' similarities based on user's interest sequence, and improves the performance of collaborative filtering recommendation. 
\citeauthor{parsana2018improving}~\shortcite{parsana2018improving} improves native ads CTR prediction by using large scale event embedding and attentional output of recurrent networks.
ATRank~\cite{zhou2017atrank} uses attention-based sequential framework to model heterogeneous behaviors. 
Compared to sequence-independent approaches,  these methods can significantly improve the prediction accuracy. 

However, these traditional RNN based models have some problems. On the one hand, most of them regard hidden states of sequential structure as latent interests directly, while these hidden states lack special supervision for interest representation. On the other hand, most existing RNN based models deal with all dependencies between adjacent behaviors successively and equally. As we know, not all user's behaviors are strictly dependent on each adjacent behavior. Each user has diverse interests,  and each interest has its own evolving track. For any target item, these models can only obtain one fixed interest evolving track,  so these models can be disturbed by interest drifting.

In order to push hidden states of sequential structure to represent latent interests effectively, extra supervision for hidden states should be introdueced. 
DARNN~\cite{ren2018learning} uses click-level sequential prediction, which models the click action at each time when each ad is shown to the
user. Besides click action, ranking information can be further introduced. In recommendation system, ranking loss has been widely used for ranking task~\cite{rendle2009bpr,hidasi2017recurrent}. Similar to these ranking losses, we propose an auxiliary loss for interest learning.  At each step, the auxiliary loss uses consecutive clicked item against non-clicked item to supervise the learning of interest representation.

For capturing interest evolving process that is related to target item, we need more flexible sequential learning structure. In the area of question answering~(QA), DMN+~\cite{xiong2016dynamic} uses attention based GRU~(AGRU) to push the attention mechanism to be sensitive to both the position and ordering of the inputs facts. In AGRU, the vector of the update gate is replaced by the scalar of attention score simply. This replacement neglects the difference between all dimensions of update gates, which contains rich information transmitted from previous sequence. Inspired by the novel sequential structure used in QA, we propose GRU with attentional gate~(AUGRU) to activate relative interests during interest evolving. Different from AGRU, attention score in AUGRU acts on the information computed from update gate. The combination of update gate and attention score pushes the process of evolving more specifically and sensitively.
\section{Deep Interest Evolution Network}
In this section, we introduce Deep Interest Evolution Network~(DIEN) in detail. First, we review the basic Deep CTR model, named BaseModel. Then we show the overall structure of DIEN and introduce the techniques that are used for capturing interests and modeling interest evolution process.
\subsection{Review of BaseModel}
The BaseModel is introduced from the aspects of feature representation, model structure and loss function.
\subsubsection{Feature Representation}
In our online display system, we use four categories of feature: {\it User Profile}, {\it User Behavior}, {\it Ad} and {\it Context}. 
It is notable that the ad is also item. For generation, we call the ad as the target item in this paper. 
Each category of feature has several fields, {\it User Profile}'s fields are {\it gender}, {\it age} and so on; 
The fields of {\it User Behavior}'s are the list of user {\it visited goods id};
{\it Ad}'s  fields are {\it ad\_id}, {\it shop\_id} and so on;
{\it Context}'s fields are {\it time} and so on.
Feature in each field can be encoded into one-hot vector, e.g., the female feature in the category of {\it User Profile} are encoded as $[0,1]$. The concat of different fields' one-hot vectors from {\it User Profile}, {\it User Behavior}, {\it Ad} and {\it Context} form $\mathbf{x}_p,\mathbf{x}_b,\mathbf{x}_a,\mathbf{x}_c$, respectively. 	
In sequential CTR model, it's remarkable that each field contains a list of behaviors, and each behavior corresponds to a one-hot vector, which can be represented by 
$\mathbf{x}_b=[\mathbf{b}_1;\mathbf{b}_2;\cdots;\mathbf{b}_T]\in\mathbb{R}^{K \times T},
\mathbf{b}_t\in\{0,1\}^{K}, $
where  $\mathbf{b}_t$ is encoded as one-hot vector and represents $t$-th behavior, $T$ is the number of user's history behaviors, $K$ is the total number of goods that user can click. 

\subsubsection{The Structure of BaseModel}
Most deep CTR models are built on the basic structure of embedding \& MLR. The basic structure is composed of several parts:
\begin{itemize}
	\item{{\it \bf{Embedding}} Embedding is the common operation that transforms the large scale sparse feature into low-dimensional dense feature.
		In the embedding layer, each field of feature is corresponding to one embedding matrix, e.g., the embedding matrix of visited goods can be represented by 
		$\mathrm{E}_{goods}=[\mathbf{m}_1;\mathbf{m}_2;\cdots;\mathbf{m}_K] \in\mathbb{R}^{n_E \times K},$
		where $\mathbf{m}_j\in\mathbb{R}^{n_E}$ represents a embedding vector with dimension $n_E$. 
		Especially, for behavior feature $\mathbf{b}_t$, if $\mathbf{b}_t[j_t] = 1$, then its corresponding embedding vector is $\mathbf{m}_{j_t}$, and the ordered embedding vector list of behaviors for one user can be represented by 
		$\mathbf{e}_b=[\mathbf{m}_{j_1}; \mathbf{m}_{j_2}; \cdots, \mathbf{m}_{j_T}].$
		Similarly, $\mathbf{e}_a$ represents the concatenated embedding vectors of fields in the category of advertisement.}
	\item{ {\it \bf{Multilayer Perceptron}}~(MLP) First, the embedding vectors from one category are fed into pooling operation. Then all these pooling vectors from different categories are concatenated. At last, the concatenated vector is fed into the following MLP for final prediction.}
\end{itemize}
\subsubsection{Loss Function}
The widely used loss function in deep CTR models is negative log-likelihood function, which uses the label of target item to supervise overall prediction:
\begin{small}
	\begin{equation}
	\label{eq:tar}
	L_{target} = - \frac{1}{N} \sum_{(\mathbf{x},y)\in\mathcal{D}}^N (y\log p(\mathbf{x}) + (1-y)\log(1-p(\mathbf{x}))),
	\end{equation}
\end{small}
where $\mathbf{x}=[\mathbf{x}_p,\mathbf{x}_a,\mathbf{x}_c,\mathbf{x}_b]\in\mathcal{D}$, $\mathcal{D}$ is the training set of size $N$. $y\in\{0,1\}$ represents whether the user clicks target item. $p(\mathbf{x})$ is the output of network, which is the predicted probability that the user clicks target item.
\subsection{Deep Interest Evolution Network}
\begin{figure*}[htp]
	\centering
	\includegraphics[height=3.1in, width=7.0in]{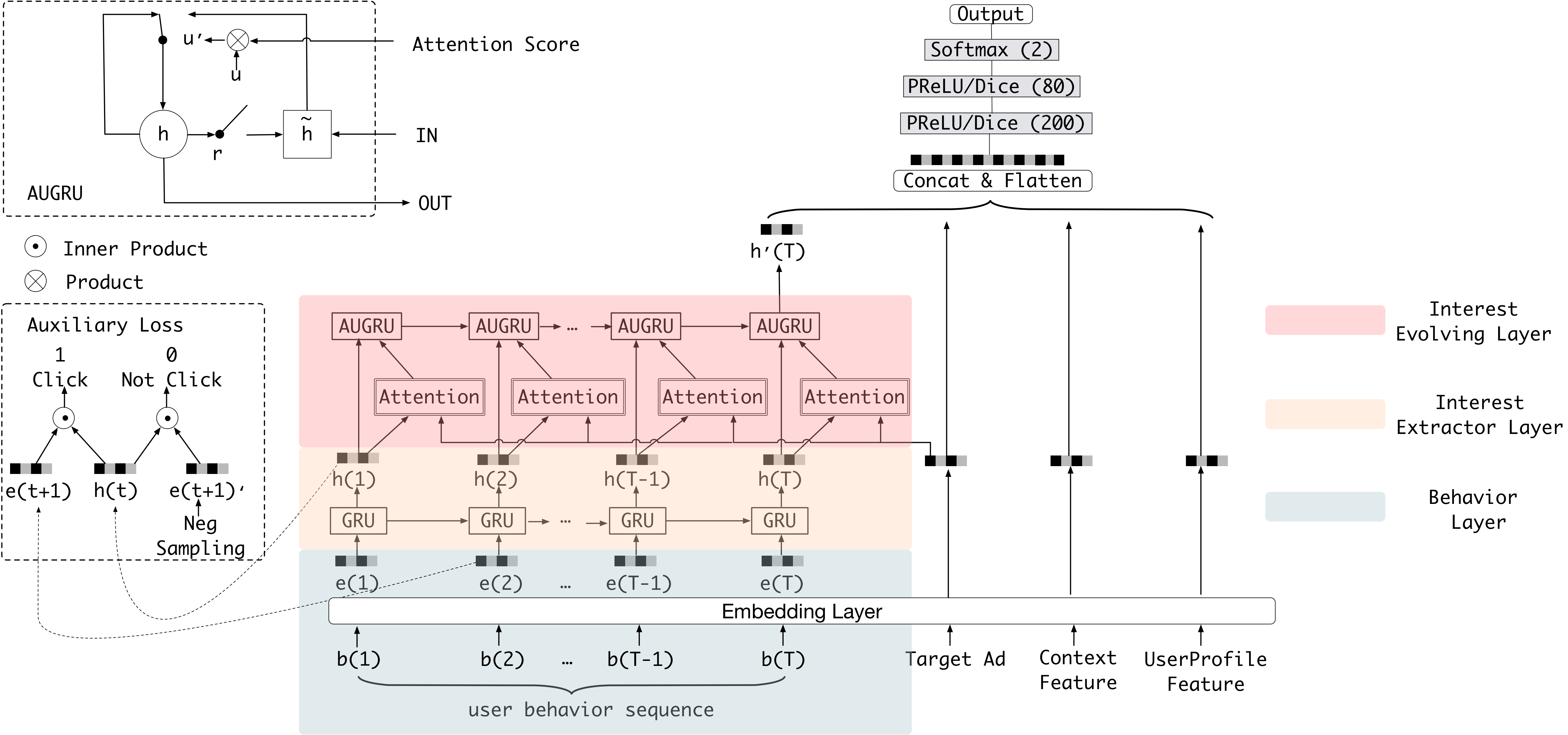}
	\caption{The  structure of DIEN.  At the behavior layer, behaviors are sorted by time, the embedding layer transforms the one-hot representation $\mathbf{b}[t]$ to embedding vector $\mathbf{e}[t]$. Then interest extractor layer extracts each interest state $\mathbf{h}[t]$ with the help of auxiliary loss. At interest evolving layer, AUGRU models the interest evolving process that is relative to target item. The final interest state $\mathbf{h}'[T]$ and embedding vectors of remaining feature are concatenated, and fed into MLR for final CTR prediction. }
	\label{fig:DIEN}
\end{figure*}

Different from sponsored search, in many e-commerce platforms like online display advertisement, users do not show their intention clearly, so capturing user interest and their dynamics is important for CTR prediction. DIEN devotes to capture user interest and models interest evolving process. 
As shown in Fig.~\ref{fig:DIEN}, DIEN is composed by several parts. First, all categories of features are transformed by embedding layer. Next, DIEN takes two steps to capture interest evolving:
interest extractor layer extracts interest sequence based on behavior sequence;  interest evolving layer models interest evolving process that is relative to target item. 
Then final interest's representation and embedding vectors of ad, user profile, context are concatenated. The concatenated vector is fed into MLP for final prediction. 
In the remaining of this section, we will introduce two core modules of DIEN in detail. 
\subsubsection{Interest Extractor Layer}
In e-commerce system, user behavior is the carrier of latent interest, and interest will change after user takes one behavior. At the interest extractor layer, we extract a series of interest states from sequential user behaviors. 

The click behaviors of user in e-commerce system are rich, where the length of history behavior sequence is long even in a short period of time, like two weeks.
For the balance between efficiency and performance, we take GRU to model the dependency between behaviors, where the input of GRU is ordered behaviors by their occur time. 
GRU overcomes the vanishing gradients problem of RNN and is faster than LSTM~\cite{hochreiter1997long}, which is suitable for e-commerce system.
The formulations of GRU are listed as follows:
\begin{small}
	\begin{eqnarray}\label{eq:gru1}
	&&\mathbf{u}_t =  \sigma(W^u\mathbf{i}_t  + U^u\mathbf{h}_{t-1}+\mathbf{b}^u),\\
	&&\mathbf{r}_t =  \sigma( W^r\mathbf{i}_t +  U^r\mathbf{h}_{t-1}+\mathbf{b}^r),\\
	&&\mathbf{\tilde{h}}_{t} = \tanh(W^h\mathbf{i}_t +  \mathbf{r}_t\circ U^h\mathbf{h}_{t-1}+\mathbf{b}^h),\\
	&&\mathbf{h}_{t} = (\mathbf{1} - \mathbf{u}_t) \circ \mathbf{h}_{t-1} + \mathbf{u}_t \circ \mathbf{\tilde{h}}_{t},\label{eq:gru4}
	\end{eqnarray}
\end{small}
where $\sigma$ is the sigmoid activation function, $\circ$ is  element-wise product, $W^u, W^r, W^h\in\mathbb{R}^{n_H \times n_I}$, 
$U^z, U^r, U^h\in{n_H\times n_H}$, $n_H$ is the hidden size, and $n_I$ is the input size. 
$\mathbf{i}_t$ is the input of GRU, $\mathbf{i}_t=\mathbf{e}_b[t]$ represents the $t$-th behavior that the user taken,  $\mathbf{h}_t$ is the $t$-th hidden states.

However, the hidden state $\mathbf{h}_t$ which only captures the dependency between behaviors can not represent interest effectively. As the click behavior of target item is triggered by final interest, the label used in $L_{target}$ only contains the ground truth that supervises final interest's prediction, while history state $\mathbf{h}_t~(t<T)$ can't obtain proper supervision. As we all know, interest state at each step leads to consecutive behavior directly.
So we propose auxiliary loss, which uses behavior $\mathbf{b}_{t +1}$ to supervise the learning of interest state $\mathbf{h}_t$. 
Besides using the real next behavior as positive instance, we also use negative instance that samples from item set except the clicked item.
There are $N$ pairs of behavior embedding sequence: $\{\mathbf{e}^i_b,\hat{\mathbf{e}}^i_b\}\in\mathcal{D_B}, i\in{1,2,\cdots,N}$, where $\mathbf{e}^i_b\in\mathbb{R}^{T\times n_E}$ represents the clicked behavior sequence, and $\hat{\mathbf{e}}^i_b\in\mathbb{R}^{T\times n_E}$ represent the negative sample sequence. $T$ is the number of history behaviors, $n_E$ is the dimension of embedding, $\mathbf{e}^i_b[t]\in\mathcal{G}$ represents the $t$-th item's embedding vector that user $i$ click, $\mathcal{G}$ is the whole item set.
$\hat{\mathbf{e}}^i_b[t]\in\mathcal{G}-\mathbf{e}^i_b[t]$ represents the embedding of item that samples from the item set except the item clicked by user $i$ at $t$-th step. Auxiliary loss can be formulated as:
\begin{small}
	\begin{align}\label{eq:aux}
	L_{aux}=-&\frac{1}{N}(\sum_{i=1}^N\sum_{t}\log\sigma(\mathbf{h}^i_t,\mathbf{e}^i_b[t+1])\\ \notag
	&+ \log(1-\sigma(\mathbf{h}^i_t,\hat{\mathbf{e}}^i_b[t+1]) )),
	\end{align}
\end{small}
where $\sigma(\mathbf{x_1}, \mathbf{x_2}) = \frac{1}{1+\exp(-[\mathbf{x}_1,\mathbf{x}_2])}$ is sigmoid activation function, $\mathbf{h}^i_t$ represents the $t$-th hidden state of GRU for user $i$. 
The global loss we use in our CTR model is:
\begin{small}
	\begin{equation}
	L = L_{target} + \alpha * L_{aux},
	\end{equation}
\end{small}
where $\alpha$ is the hyper-parameter which balances the interest representation and CTR prediction.

With the help of auxiliary loss, each hidden state $\mathbf{h}_t$ is expressive enough to represent interest state after user takes behavior $\mathbf{i}_t$. The concat of all $T$ interest points 
$[\mathbf{h}_1, \mathbf{h}_2, \cdots, \mathbf{h}_T]$ composes the interest sequence that interest evolving layer can model interest evolving on. 

Overall, the introduction of auxiliary loss has several advantages: 
from the aspect of interest learning, the introduction of auxiliary loss helps each hidden state of GRU represent interest expressively. 
As for the optimization of GRU, auxiliary loss reduces the difficulty of back propagation when GRU models long history behavior sequence. 
Last but not the least, auxiliary loss gives more semantic information for the learning of embedding layer, which leads to a better embedding matrix.

\subsubsection{Interest Evolving Layer}
As the joint influence from external environment and internal cognition, different kinds of user interests are evolving over time. Using the interest on clothes as an example, with the changing of population trend and user taste, user's preference for clothes evolves. 
The evolving process of the user interest on clothes will directly decides CTR prediction for candidate clothes. 
The advantages of modeling the evolving process is as follows:
\begin{itemize}
	\item Interest evolving module could supply the representation of final interest with more relative history information;
	\item It is better to predict the CTR of  target item by following the interest evolution trend.
\end{itemize}
Notably, interest shows two characteristics during evolving:
\begin{itemize}
	\item As the diversity of interests, interest can drift. The effect of interest drifting on behaviors is that user may interest in kinds of books during a period of time, and need clothes in another time. 
	\item Though interests may affect each other, each interest has its own evolving process, e.g. the evolving process of books and clothes is almost individually. We only concerns the evolving process that is relative to target item.
\end{itemize}
In the first stage, with the help of auxiliary loss, we has obtained expressive representation of interest sequence.
By analyzing the characteristics of interest evolving, 
we combine the local activation ability of attention mechanism and sequential learning ability from GRU to model interest evolving. The local activation during each step of GRU can intensify relative interest's effect, and weaken the disturbance from interest drifting, which is helpful for modeling interest evolving process that relative to target item.  

Similar to the formulations shown in Eq.~(\ref{eq:gru1}-\ref{eq:gru4}), we use $\mathbf{i}_t'$, $\mathbf{h}_t'$ to represent the input and hidden state in interest evolving module, where the input of second GRU is the corresponding interest state at Interest Extractor Layer: $\mathbf{i}_t'=\mathbf{h}_t$. The last hidden state $\mathbf{h}_T'$ represents final interest state.

And the attention function we used in interest evolving module can be formulated as:
\begin{small}
	\begin{equation}
	a_t = \frac{\exp(\mathbf{h}_tW\mathbf{e}_a)}{\sum_{j=1}^T\exp(\mathbf{h}_jW\mathbf{e}_a)},
	\end{equation}
\end{small}
where $\mathbf{e}_a$ is the concat of embedding vectors from fields in category ad, $W\in\mathbb{R}^{n_H\times n_A}$, $n_H$ is the dimension of hidden state and $n_A$ is the dimension of advertisement's embedding vector. Attention score can reflect the relationship between advertisement $\mathbf{e}_a$ and input $\mathbf{h}_t$, and strong relativeness leads to a large attention score.

Next, we will introduce several approaches that combine the mechanism of attention and GRU to model the process of interest evolution.
\begin{itemize}
	\item{{\bf GRU with attentional input~(AIGRU)}
		In order to  activate relative interests during interest evolution, we propose a naive method, named GRU with attentional input~(AIGRU). AIGRU uses attention score to affect the input of interest evolving layer.
		As shown in Eq.~(\ref{eq:aigru}): 
		\begin{small}
			\begin{equation}
			\label{eq:aigru}
			\mathbf{i}_t'= \mathbf{h}_t * a_t
			\end{equation}
		\end{small}
		Where $\mathbf{h}_t$ is the $t$-th hidden state of GRU at interest extractor layer, $\mathbf{i}_t'$ is the input of the second GRU which is for interest evolving, and $*$ means scalar-vector product.
		
		In AIGRU, the scale of less related interest can be reduced by the attention score. Ideally, the input value of less related interest can be reduced to zero. However, AIGRU works not very well. Because even zero input can also change the hidden state of GRU, so the less relative interests also affect the learning of interest evolving.}
	
	\item{{\bf Attention based GRU(AGRU)}
		In the area of question answering~\cite{xiong2016dynamic}, attention based GRU~(AGRU) is firstly proposed. After modifying the GRU architecture by embedding information from the attention mechanism, AGRU can extract key information in complex queries effectively. 
		Inspired by the question answering system, we transfer the using of AGRU from extracting key information in query to capture relative interest during interest evolving novelly. 
		In detail, AGRU uses the attention score to replace the update gate of GRU, and changes the hidden state directly. Formally:
		\begin{small}
			\begin{equation}
			\mathbf{h}_{t}' = (1 - a_t) * \mathbf{h}_{t-1}' + a_t * \tilde{\mathbf{{h}}}_{t}',
			\end{equation}
		\end{small}
		where $\mathbf{h}_t'$, $\mathbf{h}_{t-1}'$ and $\tilde{\mathbf{h}}_t'$ are the hidden state of AGRU.
		
		In the scene of interest evolving, AGRU makes use of the attention score to control the update of hidden state directly. AGRU weakens the effect from less related interest during interest evolving. 
		The embedding of attention into GRU improves the influence of attention mechanism, and helps AGRU overcome the defects of AIGRU.}
	
	\item{{\bf GRU with attentional update gate~(AUGRU)}
		Although AGRU can use attention score to control the update of hidden state directly, it use s a scalar~(the attention score $a_t$) to replace a vector~(the update gate $u_t$), which ignores the difference of  importance among different dimensions. We propose the GRU with attentional update gate~({\sc AUGRU}) to combine attention mechanism and GRU seamlessly:
		\begin{small}
			\begin{align}
			\mathbf{\tilde{u}}_t' &= a_t * \mathbf{u}_{t}',\\\label{eq:att}
			\mathbf{h}_{t}' &= (1 - \mathbf{\tilde{u}}_t') \circ \mathbf{h}_{t-1}' + \mathbf{\tilde{u}}_t' \circ \mathbf{\tilde{h}}_{t}',
			\end{align}
		\end{small}
		where $\mathbf{u}_{t}'$ is the original update gate of AUGRU, $\mathbf{\tilde{u}}_t'$ is the attentional update gate we design for AUGRU, $\mathbf{h}_t', \mathbf{h}_{t-1}'$,  and $\tilde{\mathbf{h}}_t'$ are the hidden states of AUGRU. 
		
		In AUGRU, we keep original dimensional information of update gate, which decides the importance of each dimension. Based on the differentiated information, we use attention score $a_t$ to scale all dimensions of update gate, which results that less related interest make less effects on the hidden state. AUGRU avoids the disturbance from interest drifting more effectively, and pushes the relative interest to evolve smoothly.  
	}
\end{itemize}

\section{Experiments}
In this section, we compare DIEN with the state of the art on both public and industrial datasets. Besides, we design experiments to verify the effect of auxiliary loss and AUGRU, respectively. For observing the process of interest evolving, we show the visualization result of interest hidden states. At last, we share the results and techniques for online serving.
\subsection{Datasets}
We use both public and industrial datasets to verify the effect of DIEN.  The statistics of all datasets are shown in Table~\ref{tab:dataset}.

\begin{table}
	\centering
	\caption{The statistics of datasets}\label{tab:dataset}
	\resizebox{1\columnwidth}{!}{%
		\begin{tabular}{c r r r r }
			\addlinespace
			\toprule
			Dataset                                                         & User & Goods  & Categories & Samples \\
			\midrule
			Books                                                           & 603,668 & 367,982 & 1,600 & 603,668 \\
			Electronics                                                 & 192,403  &  63,001 & 801 & 192,403  \\  
			Industrial dataset					   & 0.8~billion & 0.82~billion & 18,006 & 7.0~billion \\
			\bottomrule
	\end{tabular}}
\end{table}
\subsubsection{public Dataset}
Amazon Dataset~\cite{mcauley2015image} is composed of product reviews and metadata from Amazon. 
We use two subsets of Amazon dataset: Books and Electronics, to verify the effect of DIEN. 
In these datasets, we regard reviews as behaviors, and sort the reviews from one user by time. Assuming there are $T$ behaviors of user $u$, our purpose is to use the $T-1$ behaviors to predict whether user $u$ will write reviews that shown in $T$-th review. 

\subsubsection{Industrial Dataset}
Industrial dataset is constructed by impression and click logs from our online display advertising system. 
For training set, we take the ads that clicked at last $49$ days as the target item. Each target item and its corresponding click behaviors construct one instance. Using one target item $a$ as example, we set the day that $a$ is clicked as the last day, the behaviors that this user takes in previous $14$ days as history behaviors. Similarly, the target item in test set is choose from the following one day, and the behaviors are built as same as training data. 

\subsection{Compared Methods}
We compare DIEN with some mainstream CTR prediction methods:
\begin{itemize}
	\item{{\it \textbf{BaseModel}}} BaseModel takes the same setting of embedding and MLR with DIEN, and uses sum pooling operation to integrate behavior embeddings.
	\item{\it \textbf{Wide\&Deep}}~\cite{cheng2016wide} Wide \& Deep consists of  two parts: its deep model is the same as Base Model, and its wide model is a linear model. 
	\item{\it \textbf{PNN}}~\cite{qu2016product} PNN uses a product layer to capture interactive patterns between interfield categories.
	\item{\it \textbf{DIN}}~\cite{zhou2018deep} DIN uses the mechanism of attention to activate related user behaviors.
	\item{\it \textbf{Two layer GRU Attention}} Similar to \cite{parsana2018improving}, we use two layer GRU to model sequential behaviors, and takes an attention layer to active relative behaviors.
\end{itemize}
\subsection{\textbf{Results on Public Datasets}}
Overall, as shown in Fig.~\ref{fig:DIEN}, the structure of DIEN consists GRU, AUGRU and auxiliary loss and other normal components. In public datase. Each experiment is repeated 5 times. 

\begin{table}
	\small
	\centering
	\caption{Results~(AUC) on public datasets}\label{tab:public}
	\resizebox{\columnwidth}{!}{%
		\begin{tabular}{l c c}
			\addlinespace
			\toprule
			Model               & Electronics~(mean$\pm$ std) & Books~(mean $\pm$ std) \\
			\midrule
			{\it BaseModel}~\cite{zhou2018deep}    & $0.7435\pm 0.00128$  &  $0.7686\pm 0.00253$  \\
			{\it Wide\&Deep}~\cite{cheng2016wide}      & $0.7456\pm 0.00127$ & $0.7735\pm 0.00051$   \\
			{\it PNN}~\cite{qu2016product}              & $0.7543\pm 0.00101$  & $0.7799\pm 0.00181$ \\
			{\it DIN}~\cite{zhou2018deep}                   & $0.7603\pm 0.00028$  & $0.7880\pm 0.00216$ \\
			{\it Two layer GRU Attention}                  & $0.7605\pm 0.00059$  & $0.7890\pm 0.00268$ \\
			{\it DIEN}            & $\bm{0.7792\pm 0.00243}$  & $\bm{0.8453\pm 0.00476}$ \\
			\bottomrule
	\end{tabular}}
\end{table}
From Table~\ref{tab:public}, we can find {\it Wide \& Deep} with manually designed features performs not well, while automative interaction between features~({\it PNN}) can improve the performance of {\it BaseModel}.
At the same time, the models aiming to capture interests can improve AUC obviously: {\it DIN} actives the interests that relative to ad, 
{\it Two layer GRU attention} further activates relevant interests in interest sequence, all these explorations obtain positive feedback. 
DIEN not only captures sequential interests more effectively, but also models the interest evolving process that is relative to target item. The modeling for interest evolving helps DIEN obtain better interest representation, and capture dynamic of interests precisely, which improves the performance largely. 
\subsection{Results on Industrial Dataset}
We further conduct experiments on the dataset of real display advertisement. 
There are 6 FCN layers used in industrial dataset, the dimensions area 600, 400, 300, 200, 80, 2, respectively, the max length of history behaviors is set as $50$.

As shown in Table \ref{tab:industrial}, {\it Wide \& Deep} and {\it PNN} obtain better performance than {\it BaseModel}. 
Different from only one category of goods in Amazon dataset, the dataset of online advertisement contains all kinds of goods at the same time. Based on this characteristic, attention-based methods improve the performance largely, like {\it DIN}. 
DIEN captures the interest evolving process that is relative to target item, and obtains best performance. 
\begin{table}[!htbp]
	\centering
	\caption{Results~(AUC) on industrial dataset}\label{tab:industrial}
	\resizebox{0.28\textwidth}{!}{%
		\begin{tabular}{l c c}
			\addlinespace
			\toprule
			Model           &AUC      \\
			\midrule
			{\it BaseModel}~\cite{zhou2018deep}  &0.6350 \\
			{\it Wide\&Deep}~\cite{cheng2016wide}& 0.6362       \\
			{\it PNN}~\cite{qu2016product}            & 0.6353  \\
			{\it DIN}~\cite{zhou2018deep}               & 0.6428 \\
			{\it Two layer GRU Attention}  & 0.6457\\
			{\it BaseModel +  GRU + AUGRU}      &0.6493\\
			{\it DIEN}            & $\bm{0.6541}$       \\
			\bottomrule
	\end{tabular}}
\end{table}
\subsection{Application Study}
In this section, we will show the effect of AUGRU and auxiliary loss, respectively. 
\subsubsection{Effect of GRU with attentional update gate~(AUGRU)}\label{sess:att}
\begin{table}[!htbp]
	\small
	\centering
	\caption{Effect of AUGRU and auxiliary loss~(AUC)}\label{tab:att}
	\resizebox{1\columnwidth}{!}{%
		\begin{tabular}{l c c}
			\addlinespace
			\toprule
			Model                                                             & Electronics~(mean $\pm$ std) & Books~(mean $\pm$ std) \\
			\midrule
			{\it BaseModel}                                                        & $0.7435\pm 0.00128$  &  $0.7686\pm 0.00253$  \\
			{\it Two layer GRU attention}                                   & $0.7605\pm 0.00059$  & $0.7890\pm 0.00268$\\
			{\it BaseModel +  GRU + AIGRU}                           & $0.7606\pm 0.00061$  & $0.7892\pm 0.00222$\\
			{\it BaseModel +  GRU + AGRU}                           & $0.7628\pm 0.00015$  & $0.7890\pm 0.00268$ \\
			{\it BaseModel +  GRU + AUGRU}               & $0.7640\pm 0.00073$  & $0.7911\pm 0.00150$  \\
			{\it DIEN}                & $\bm{0.7792\pm 0.00243}$  & $\bm{0.8453\pm 0.00476}$  \\
			\bottomrule
	\end{tabular}}
\end{table}
Table~\ref{tab:att} shows the results of different methods for interest evolving. 
Compared to {\it BaseMode}, {\it Two layer GRU Attention} obtains improvement, while the lack for modeling evolution limits its ability.
{\it AIGRU} takes the basic idea to model evolving process, 
though it has advances, the splitting of attention and evolving lost information during interest evolving.
{\it AGRU} further tries to fuse attention and evolution, as we proposed previously, its attention in GRU can't make fully use the resource of update gate.
{\it AUGRU} obtains obvious improvements, which reflects it fuses the attention mechanism and sequential learning ideally, and captures the evolving process of relative interests effectively.          
\subsubsection{Effect of auxiliary loss}
Based on the model that obtained with AUGRU, we further explore the effect of auxiliary loss. 
In the public datasets, the negative instance used in the auxiliary loss is randomly sampled from item set except the item shown in corresponding review. 
As for industrial dataset, the ads shown while not been clicked are as negative instances. 
\begin{figure}[!htb]
	\centering
	\begin{subfigure}[t]{0.23\textwidth}
		\centering
		\includegraphics[width=4.1cm]{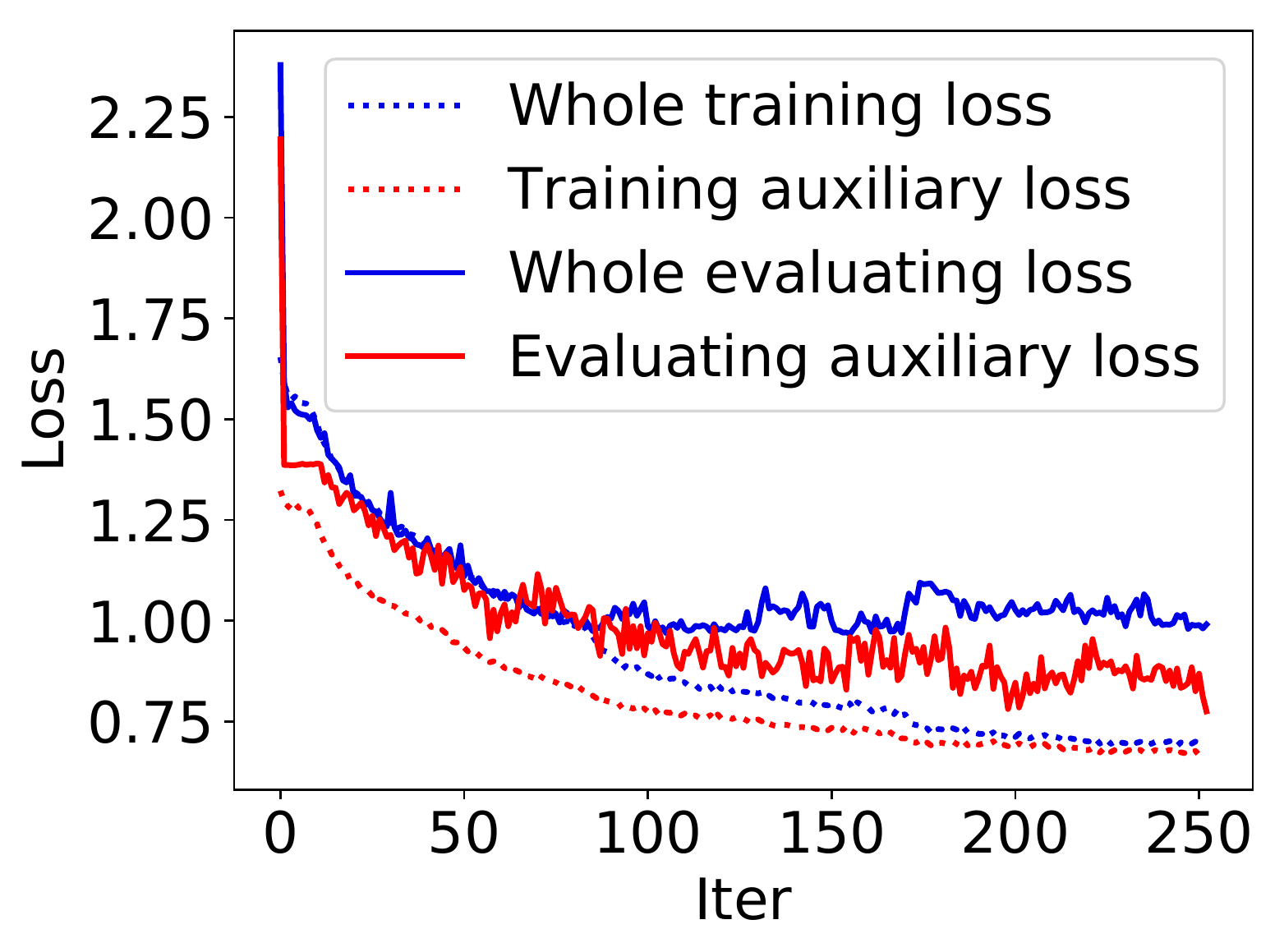}
		\subcaption{Books}
	\end{subfigure}
	\begin{subfigure}[t]{0.23\textwidth}
		\centering
		\includegraphics[width=4.1cm]{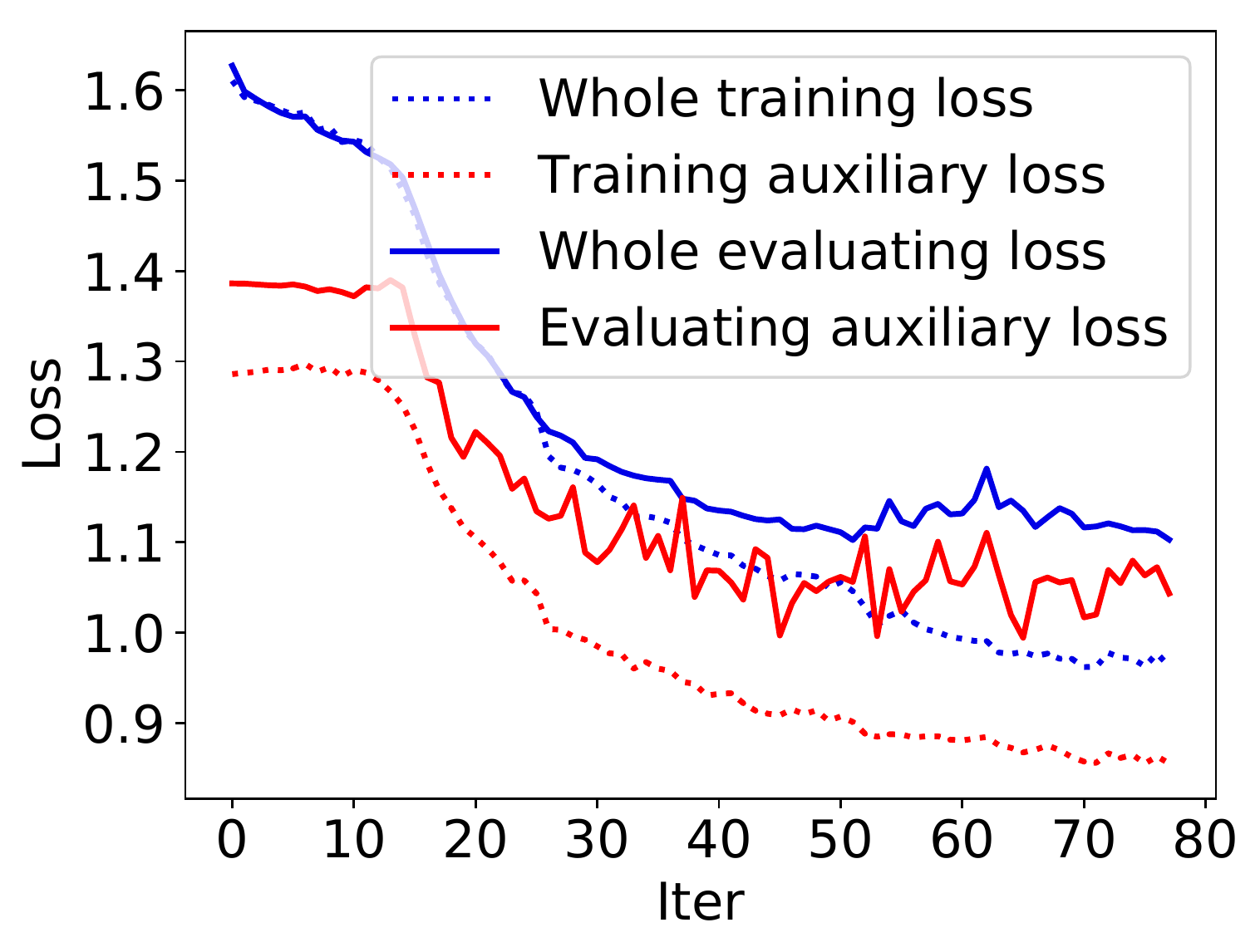}
		\subcaption{Electronics}
	\end{subfigure}
	\caption{Learning curves on public datasets. $\alpha$ is set as $1$.  }
	\label{fig:loss}
\end{figure}
As shown in Fig.~\ref{fig:loss}, the loss of both whole loss $L$ and auxiliary loss $L\_aux$ keep similar descend trend, which means both global loss for CTR prediction and auxiliary loss for interest representation make effect. 

As shown in Table~\ref{tab:att}, auxiliary loss bring great improvements for both public datasets, it reflects the importance of supervision information for the learning of sequential interests and  embedding representation.  
For industrial dataset shown in Table~\ref{tab:industrial}, model with auxiliary loss improves performance further. However, we can see that the improvement is not as obvious as that in public dataset. The difference derives from several aspects.
First, for industrial dataset, it has a large number of instances to learn the embedding layer, which makes it earn less from auxiliary loss. Second, different from all items from one category in amazon dataset, the behaviors in industrial dataset are clicked goods from all scenes and all catagories in our platform.  Our goal is to predict CTR for ad in one scene. The supervision information from auxiliary loss may be heterogeneous from the target item, so the effect of auxiliary loss for the industrial dataset may be less for public datasets, while the effect of AUGRU is magnified.
\subsection{Visualization of Interest Evolution}
The dynamic of hidden states in AUGRU can reflect the evolving process of interest. In this section, we visualize these hidden states to explore the effect of different target item for interest evolution. 

\begin{figure}[!htb]
	\centering
	\begin{subfigure}[t]{0.4\textwidth}
		\centering
		\includegraphics[width=5.5cm]{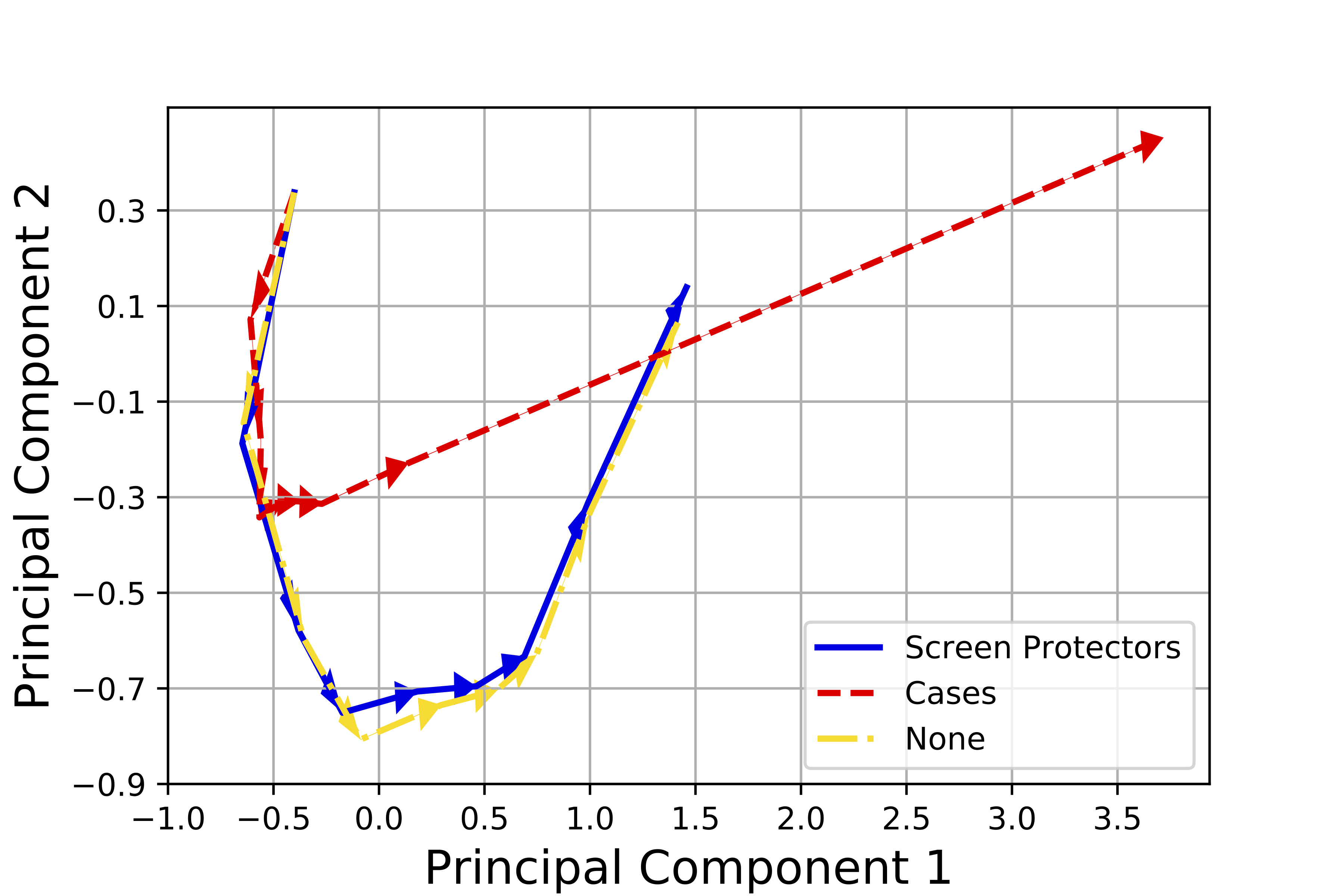}
		\subcaption{Visualization of hidden states in AUGRU}
	\end{subfigure}
	\begin{subfigure}[t]{0.4\textwidth}
		\centering
		\includegraphics[height=1.5in, width=5.3cm]{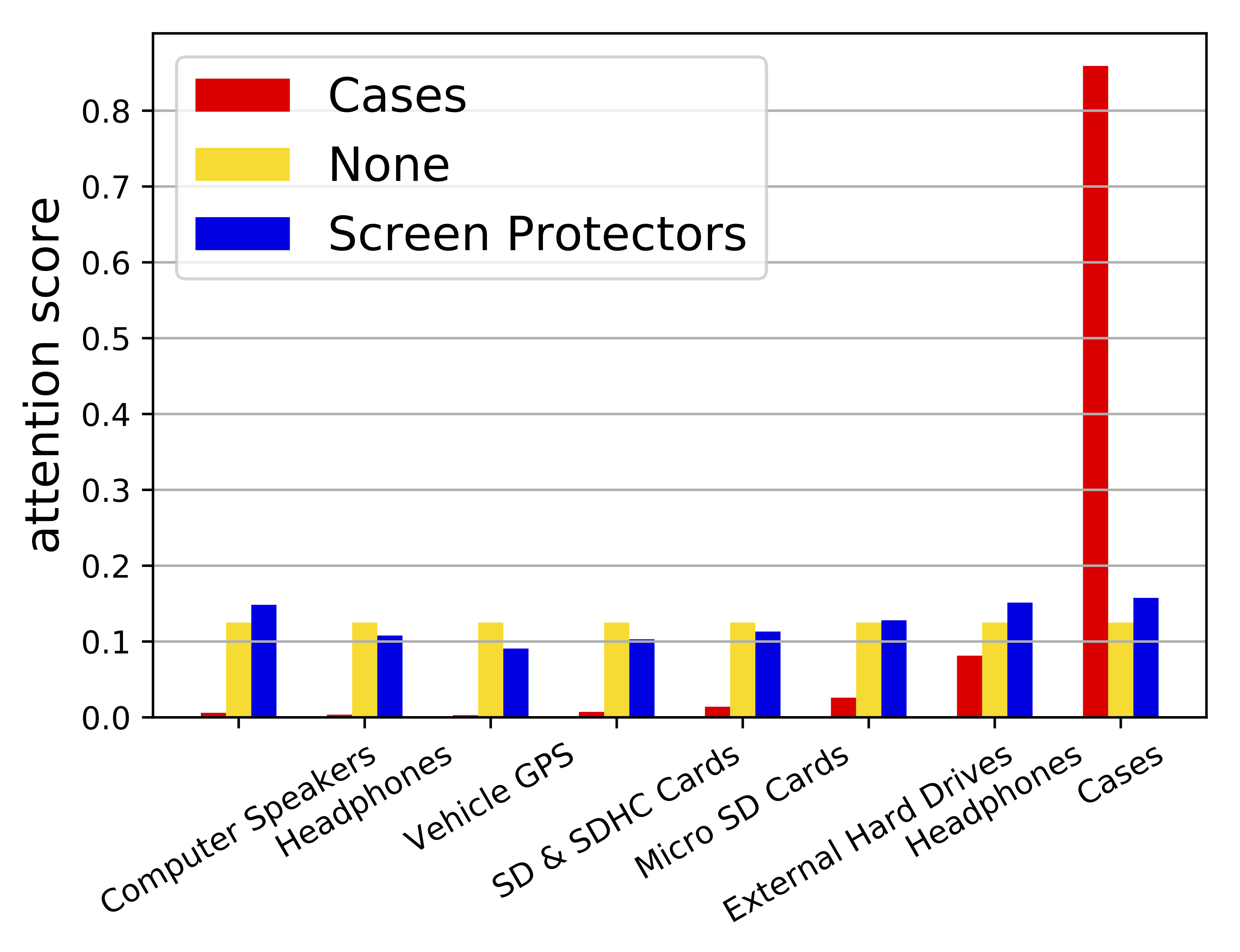}
		\subcaption{Attention score of different history behaviors}
	\end{subfigure}
	\caption{Visualization of interest evolution, (a) The hidden states of AUGRU reduced by PCA into two dimensions.  Different curves shows the same history behaviors are activated by different target items. {\it None} means interest evolving is not effected by target item. (b) Faced with different target item, attention scores of all history behaviors are shown.}\label{fig:att}
\end{figure}

The selective history behaviors are from category {\it Computer Speakers, Headphones, Vehicle GPS, SD \& SDHC Cards, Micro SD Cards, External Hard Drives, Headphones, Cases}, successively.
The hidden states in AUGRU are projected into a two dimension space by principal component analysis~(PCA)~\cite{wold1987principal}. The projected hidden states are linked in order. The moving routes of hidden states  activated by different target item are shown in Fig.~\ref{fig:att}(a). The yellow curve which is with {\it None} target represents the attention score used in eq. (\ref{eq:att}) are equal, that is the evolution of interest are not effected by target item. The blue curve shows the hidden states are activated by item from category {\it Screen Protectors}, which is less related to all history behaviors, so it shows similar route to yellow curve. The red curve shows the hidden states are activated by item from category {\it Cases}, the target item is strong related to the last behavior, which moves a long step shown in Fig. \ref{fig:att}(a). Correspondingly, the last behavior obtains a large attention score showed in  Fig. \ref{fig:att}(b).  
\subsection{Online Serving \& A/B testing}
\begin{table}
	\centering
	\caption{Results from Online A/B testing}\label{tab:OlineTest}
	\resizebox{0.40\textwidth}{!}{%
		\begin{tabular}{c r r r}
			\addlinespace
			\toprule
			Model & CTR Gain  & PPC Gain & eCPM Gain\\
			\midrule
			BaseModel & 0\% & 0\% & 0\% \\
			DIN~\cite{zhou2018deep} & + 8.9\% & - 2.0\% & + 6.7\%\\  
			DIEN & + 20.7\% & - 3.0\% & + 17.1\% \\
			\bottomrule
	\end{tabular}}
\end{table}
From 2018-06-07 to 2018-07-12, online A/B testing was conducted in the display advertising system of Taobao. As shown in Table~\ref{tab:OlineTest}, compared to the BaseModel, DIEN has improved CTR by 20.7\% and effective cost per mille~(eCPM) by 17.1\%. Besides, DIEN has decayed pay per click~(PPC) by 3.0\%. Now, DIEN has been deployed online and serves the main traffic, which contributes a significant business revenue growth.

It is worth noticing that online serving of DIEN is a great challenge for commercial system. 
Online system holds really high traffic in our display advertising system, which serves more than 1 million users per second at traffic peak. 
 In order to keep low latency and high throughput, we deploy several important techniques to improve serving performance: i) {\it  element parallel GRU \& kernel fusion}~\cite{WangKernel}, we fuse as many independent kernels as possible. Besides, each element of the hidden state of GRU can be calculated in parallel. ii) {\it Batching}: adjacent requests from different users are merged into one batch to take advantage of GPU. iii) {\it Model compressing with Rocket Launching}~\cite{zhou2018rocket}: we use the method proposed in ~\cite{zhou2018rocket} to train a light network, which has smaller size but performs close to the deeper and more complex one. For instance, the dimension of GRU hidden state can be compressed from 108 to 32 with the help of Rocket Launching. With the help of these techniques, latency of DIEN serving can be reduced from 38.2 ms to 6.6 ms and the QPS (Query Per Second) capacity of each worker can be improved to 360.
\section{Conclusion}
In this paper, we propose a new structure of deep network, namely Deep Interest Evolution Network~(DIEN), to model interest evolving process. DIEN improves the performance of CTR prediction largely in online advertising system. 
Specifically, we design interest extractor layer to capture interest sequence particularly, which uses auxiliary loss to provide the interest state with more supervision. 
Then we propose interest evolving layer, where DIEN uses GRU with attentional update gate~(AUGRU) to model the interest evolving process that is relative to target item. With the help of AUGRU, DIEN can overcome the disturbance from interest drifting. 
Modeling for interest evolution helps us capture interest effectively, which further improves the performance of CTR prediction.
In future, we will try to construct a more personalized interest model for CTR prediction. 
\bibliography{DEIN}
\bibliographystyle{aaai}
\end{document}